\documentclass[10pt,twocolumn,letterpaper]{article}

\usepackage[pagenumbers]{cvpr} 

\usepackage{graphicx}
\usepackage{amsmath}
\usepackage{amssymb}
\usepackage{booktabs}

\usepackage[pagebackref,breaklinks,colorlinks]{hyperref}

\usepackage[capitalize]{cleveref}
\crefname{section}{Sec.}{Secs.}
\Crefname{section}{Section}{Sections}
\Crefname{table}{Table}{Tables}
\crefname{table}{Tab.}{Tabs.}

\def\cvprPaperID{*****} 
\def\confName{CVPR}
\def\confYear{2022}

\begin{document}

\title{General Image Descriptors for Open World Image Retrieval using ViT CLIP}

\author{Marcos V. Conde$^{1}$, Ivan Aerlic$^{2}$, Simon Jégou$^{3}$\\
\\
$^{1}$ \url{H2O.ai} and Computer Vision Lab, CAIDAS, University of Würzburg, Germany\\
$^{2}$Independent researcher and Team Leader, Australia\\
$^{3}$Independent researcher, France\\
{\tt\small marcos.conde-osorio@uni-wuerzburg.de}\\
{\tt\small {\url{https://github.com/IvanAer/G-Universal-CLIP}}}
}

\maketitle

\begin{abstract}
   The Google Universal Image Embedding (GUIE) Challenge is one of the first competitions in multi-domain image representations in the wild, covering a wide distribution of objects: landmarks, artwork, food, etc. This is a fundamental computer vision problem with notable applications in image retrieval, search engines and e-commerce.
   
   In this work, we explain our 4th place solution to the GUIE Challenge, and our "bag of tricks" to fine-tune zero-shot Vision Transformers (ViT) pre-trained using CLIP.
   
\end{abstract}


\section{Introduction}
\label{sec:intro}

Image representations are a critical building block of computer vision applications~\cite{douze20212021}. Traditionally, research on image embedding learning has been conducted with a focus on per-domain models~\cite{araujo_retrieval, radenovic2018fine, google_landmark}. Generally, solutions are based on generic embedding learning techniques which are applied to different domains separately, rather than developing generic embedding models which could be applied to all domains combined.

At the Google Universal Image Embedding (\textbf{GUIE}) Challenge, the proposed models are expected to retrieve relevant index database images to a given query image (\emph{i.e.} images containing the same object as the query) considering a great variety of domains. 
Our proposed solution has real-world visual search applications, such as organizing photos, improving search engines, and visual e-commerce.


\vspace{-2mm}

\paragraph{Problem definition}
We seek for a function $\phi$ such that:

\begin{equation}
    \phi: \mathbb{R}^{H\times W\times 3} \mapsto \mathbb{R}^{64} \quad \phi (x) = \mathbf{q} \in \mathbb{R}^{64} 
\end{equation}

given an input 3-channel RGB image $x$ of dimension $H\times W$, our model $\phi$ extract a compact 64-dimensional ($64\mathcal{D}$) image descriptor or embedding $\phi (x)$.

Then the \textbf{image retrieval} task\cite{cao2020unifying, araujo2020deep, radenovic2018fine} considers an index-reference database of images $Z=\{z_1, z_2, \dots, z_n\}$, and a given a query image $x$, we calculate
\begin{equation}
    \underset{Z}{\mathrm{argmin}} \lVert \phi(x) - \phi(z_i) \rVert^{2}_{2}
\label{eq:search}
\end{equation}
finally retrieve the top-$k$ most similar images (\emph{i.e.} those that minimize the previous equation).

\paragraph{Evaluation}

Methods are evaluated according to the mean Precision at $k=5$ (abbreviated as $mP@5$):
\begin{equation}
    mP@5 = \frac{1}{Q} \sum_{q=1}^{Q} \frac{1}{min(n_q, 5)} \sum_{j=1}^{min(n_q, 5)} rel_q(j)
\label{eq:metric}
\end{equation}

where $Q$ is the number of query images, $n_q$ is the number of index images containing an object in common with the query image $q$. Note that $n_q>0$ for any query image $q$. The term $rel_q(j)$ denotes the relevance of prediction $j$ for the $q$-th query: $rel_q(j)=1$ if the $j$-th prediction is correct, and 0 otherwise.
Participants must submit a model file (\emph{e.g.} \texttt{.pt}). The model must take an image as an input, and return a float vector (\emph{i.e.} the image embedding) as the output. The challenge platform Kaggle use the submitted model to:
\begin{enumerate}
    \item Extract embeddings for the private test dataset (query and index images).
    \item Create a kNN ($k = 5$) lookup for each test sample, using the Euclidean distance between test and index embeddings. See Equation \ref{eq:search}.
    \item Score the quality of the lookups using Equation \ref{eq:metric}.
\end{enumerate}

\begin{figure}[h]
    \centering
    \includegraphics[width=\linewidth]{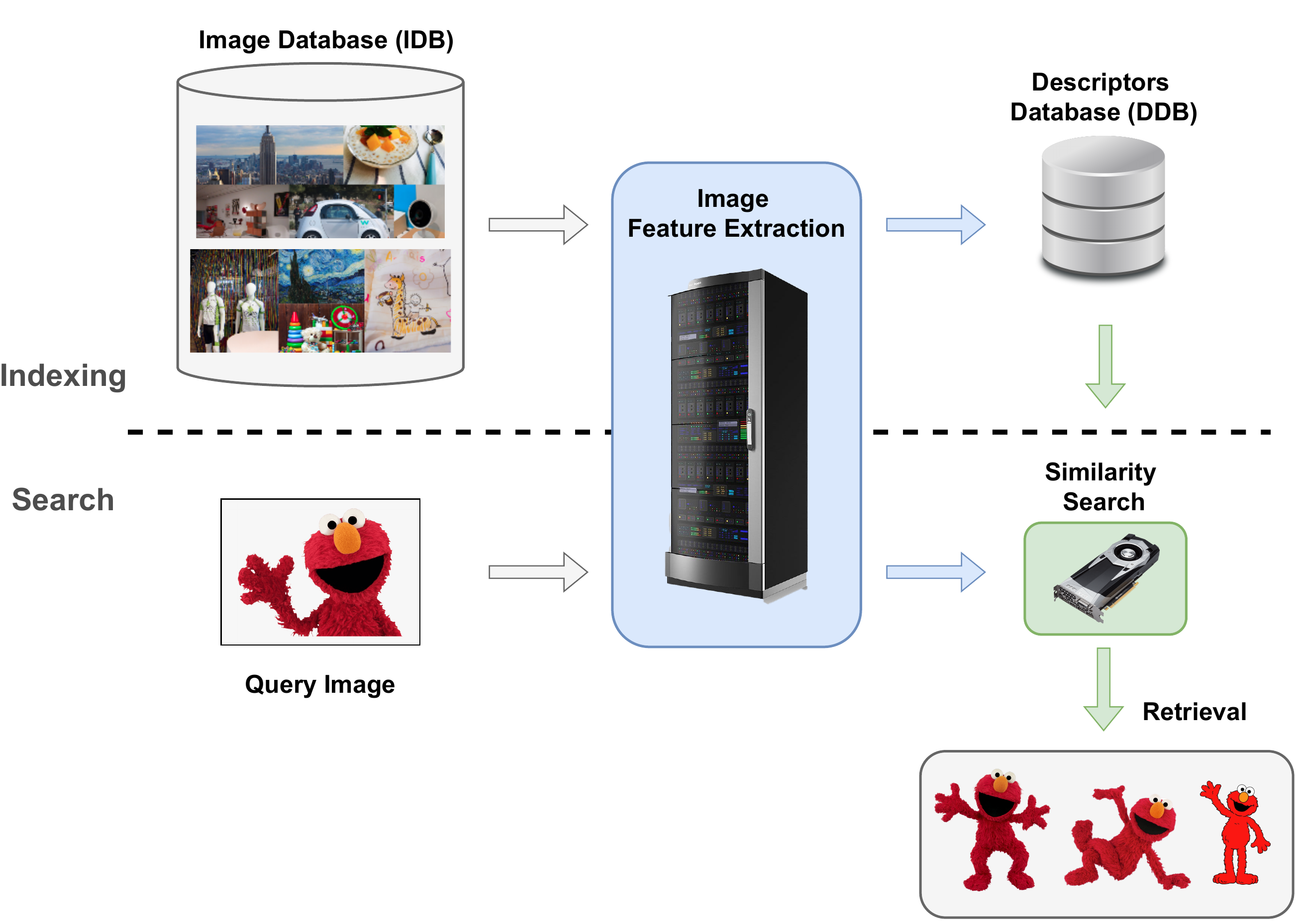}
    \caption{Example of Image Retrieval Pipeline. Inspired in Matsui \emph{et al.} CVPR 2020 Tutorial \cite{cvpr20_tutorial_image_retrieval}.}
    \label{fig:app1}
\end{figure}

In Figure~\ref{fig:app1} we provide an illustrative example of a real-world image retrieval system, similar to the one employed in this challenge for evaluating the quality of the produced image descriptors.

Note that the \textbf{evaluation dataset} is kept private. There are $\approx5000$ test query images and $\approx200.000$ index images. 

\begin{figure}[]
    \centering
    \includegraphics[width=\linewidth]{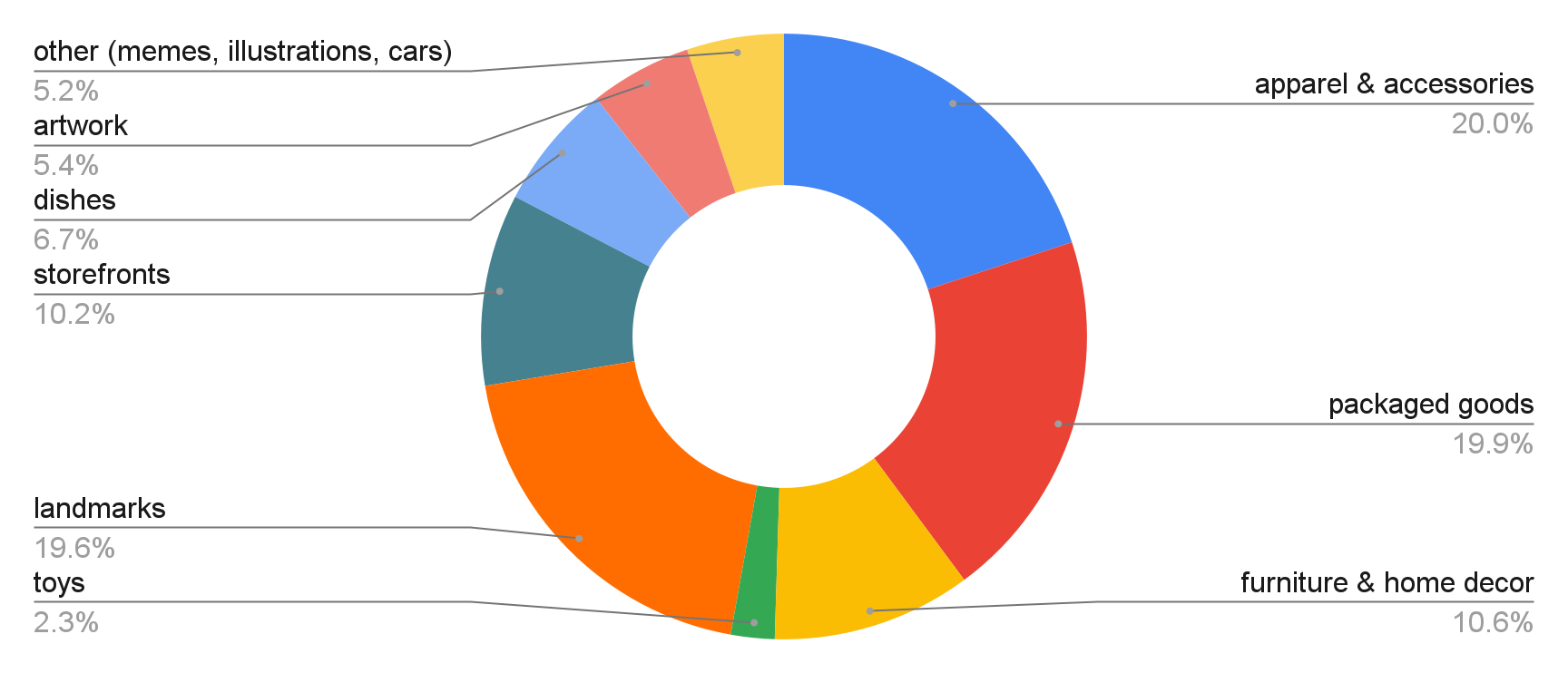}
    \caption{GUIE private evaluation dataset distribution}
    \label{fig:data-dist}
\end{figure}

In Figure \ref{fig:data-dist} we show the data distribution. The dataset images of the following types of object: apparel and accessories, packaged goods, landmarks, furniture and home decoration, storefronts, dishes, artwork, toys, memes, illustrations and cars. 
%

\section{Our approach}

\begin{figure*}[ht]
    \centering
    \includegraphics[width=0.9\linewidth]{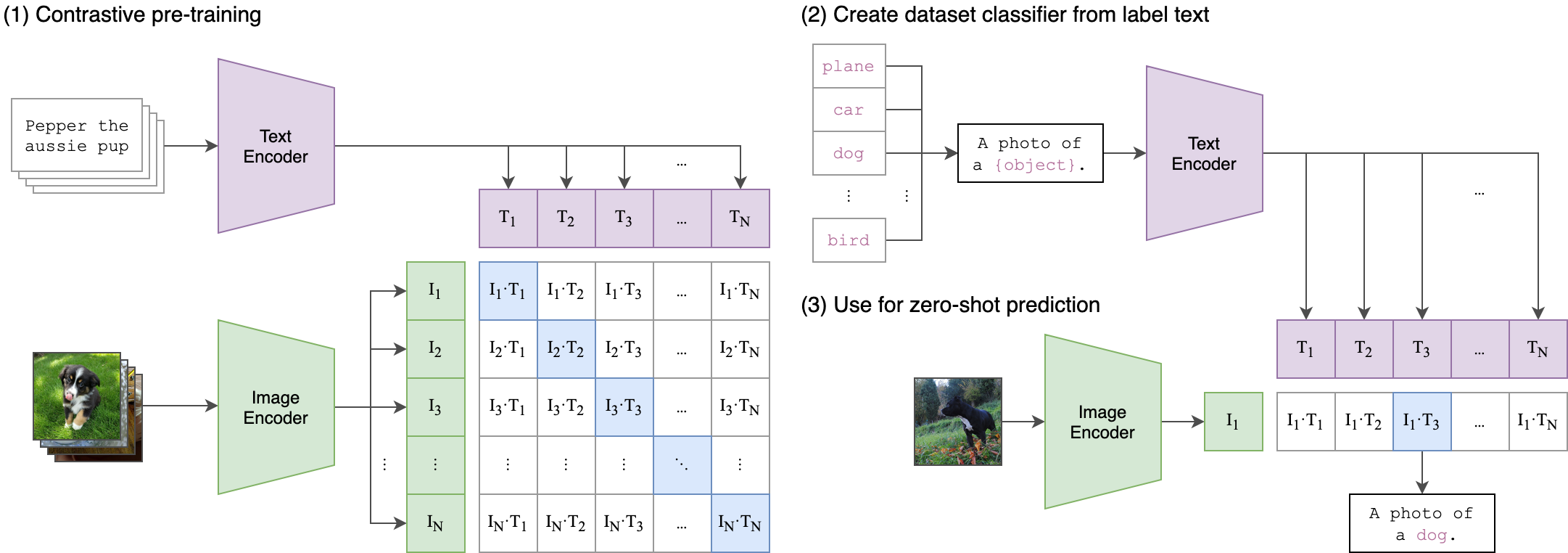}
    \caption{The original Contrastive Language–Image Pre-training (CLIP) framework proposed by Radford \emph{et al.} (OpenAI)~\cite{dosovitskiy2020vit, radford2021clip}. This allows neural networks to efficiently learn visual concepts from natural language supervision.}
    \label{fig:app2}
\end{figure*}

Zero-shot models based on CLIP~\cite{openclip, radford2021clip} have shown great success for text-image retrieval~\cite{conde2021clip, baldrati2022effective}. These models leverage images and their textual descriptions to learn rich representations. Usually the text information is encoded using a Transformer~\cite{vaswani2017attention, brown2020gpt}, a large-scale pre-trained language model. Meanwhile, the visual encoders are Vision Transformers (ViT)~\cite{dosovitskiy2020vit, conde2021exploring}, which also have shown some benefits with respect to classical CNNs at several computer vision tasks\cite{el2021training, liu2021swin}. In Figure~\ref{fig:app2} we show the CLIP framework.
These zero-shot models require billions of image-text pairs to achieve competitive performances and unfortunately, not many billion-scale image-text pair dataset had been openly available up until now. We use the recently open-sourced \textbf{OpenCLIP}~\cite{openclip} and pre-trained models on \texttt{LAION-2B}. 

As we will prove experimentally in Section~\ref{sec:zs}, the image descriptors from ViT CLIP are already a great baseline, which indicates the rich information encoded into them. In Section~\ref{sec:final} we explain our final solution consisting in a ensemble of fine-tuned models.

\subsection{Zero-shot Text-Image Alignment}
\label{sec:zs}

We use as baseline a Vision Transformer ViT-H~\cite{dosovitskiy2020vit} pre-trained on LAION-2B~\cite{radford2021clip, openclip}. This takes as input a 224px image and produces a high-dimensional encoded representation. However, the descriptor provided by ViT-H~\cite{dosovitskiy2020vit} is $1024\mathcal{D}$, therefore we perform a PCA~\cite{jegou2012pca} dimensionality reduction to obtain the desired $64\mathcal{D}$ descriptor. In this scenario we compare three different approaches:

\begin{enumerate}
    \item No PCA is performed, instead a random selection of 64 neurons (from 1024) to serve as a baseline.
    \item PCA fitted using image descriptors: We extract descriptors from our internal dataset with 10.000 images~$^1$, and fit PCA. In this scenario we require a great variety of images, which can be complicated. We believe this is the reason why performance is limited.
    \item PCA fitted using text descriptors. We will emphasize this idea as we consider this has a lot of potential.
\end{enumerate}

\begin{figure}[t]
    \centering
    \includegraphics[width=\linewidth]{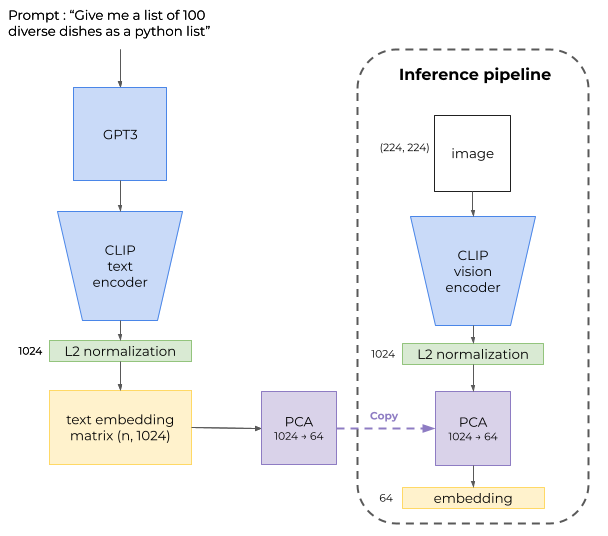}
    \caption{GUIE Zero-shot solution \cite{openclip, radford2021clip, brown2020gpt, dosovitskiy2020vit}}
    \label{fig:zero-shot}
\end{figure}

A natural way to perform PCA would have been to collect images following the distribution described on the dataset description (see Figure \ref{fig:data-dist}). Here we present an alternative that does not require any data collection: first, we generate a dataset of 3000 plausible labels using GPT3\cite{brown2020gpt} and feed them into the text encoder. The PCA is then fitted in the text embedding space and transferred to the vision encoder (ViT). The proposed baseline solution is illustrated in Figure \ref{fig:zero-shot}, it leverages CLIP ViT-H~\cite{radford2021clip} as a visual image encoder and a PCA layer fitted on data generated by GPT3\cite{brown2020gpt}. As we can see in Table~\ref{tab:lb}, this achieves 0.603 on the leaderboard compared to 0.553 for a baseline method without PCA (via random selection of 64 neurons).

\paragraph{How does it work?} The zero-shot (ZS) ViT model $\phi$ produces image descriptors which are aligned with the corresponding GPT encoder $\psi$ text descriptors due to the contrastive language-image pretraining (CLIP)\cite{radford2021clip}. For this reason, in the descriptors space $\mathcal{D}=\mathbb{R}^{64}$ an image and its text description represent the same. In other words, for an image $x$ and its text natural language description $y$:

\begin{equation}
    \lVert \phi(x) - \psi(y) \rVert \to 0
\label{eq:align}
\end{equation}

This allows us to use text descriptions instead of image descriptors to fit the PCA in such space $\mathcal{D}$. 

\begin{table}[t]
    \centering
    \begin{tabular}{l|c}
         Method & Score \\
         \hline
         ViT ZS~\cite{dosovitskiy2020vit, openclip} Random 64 neurons & 0.553 \\
         ViT ZS~\cite{dosovitskiy2020vit, openclip} + PCA on Text & \textbf{0.603} \\
         ViT ZS~\cite{dosovitskiy2020vit, openclip} + PCA on Image & 0.580 \\
         \hline
    \end{tabular}
    \caption{Summary of zero-shot (ZS) baseline methods results on the competition leaderboard (LB). Score is $mP@5$}
    \label{tab:lb}
\end{table}

\subsection{Robust Fine-tuning}
\label{sec:final}

Large pre-trained zero-shot models such as ViT CLIP\cite{dosovitskiy2020vit, radford2021clip} offer consistent accuracy across a great range of data distributions when performing zero-shot inference (\emph{i.e.} without fine-tuning on a specific dataset)\cite{wortsman2021robustwise}. CLIP-Art\cite{conde2021clip} serves as an example of fine-tuning for a specific data distribution: artwork. However, as we previously indicated the data distribution in this case is very diverse.

Although existing fine-tuning approaches substantially improve accuracy in-distribution (\emph{e.g.} Figure~\ref{fig:data-dist}), they often reduce out-of-distribution (OOD) robustness\cite{wortsman2021robustwise}.

We consider this when fine-tuning two Vision Transformers (ViT) \cite{dosovitskiy2020vit} on the challenge dataset distribution (see Figure~\ref{fig:data-dist} and Section~\ref{sec:intro}).
The two selected backbones are \texttt{ViT-L-14} and \texttt{ViT-H-14} both pre-trained on LAION-2B from OpenCLIP\cite{openclip}. The input image sizes are 336px and 224px respectively. Input images are resized to the required inputs using a bicubic interpolation.

\begin{figure}[t]
    \centering
    \includegraphics[width=\linewidth]{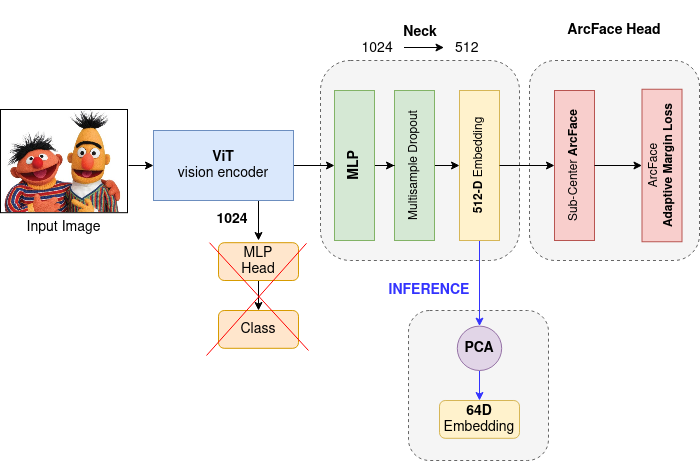}
    \caption{ViT~\cite{radford2021clip, dosovitskiy2020vit} fine-tuning using ArcFace Margin Loss~\cite{deng2019arcface}.}
    \label{fig:vit}
\end{figure}

We show the fine-tuning process and model changes in Figure \ref{fig:vit}. We reduce the $1024\mathcal{D}$ original ViT~\cite{dosovitskiy2020vit} encoding using a MLP layer with dropout, obtaining a smaller $512\mathcal{D}$ descriptor. 
Next, this is fed into a sub-center ArcFace Head~\cite{deng2019arcface} which enhances the discriminative power of the classical softmax loss to maximize class separability. 

This is also a common technique to fine-tune image retrieval models considering thousands of fine-grained classes~\cite{henkel2020supporting, conde2021exploring, douze20212021, papakipos2022results}. 
Once the model is fine-tuned, during \textbf{inference}, we obtain the final descriptor using PCA to reduce dimensions from 512 to 64.
Note that, at this point after fine-tuning, the image descriptors are no longer aligned with their related text descriptors (see Equation \ref{eq:align}), therefore we cannot fit PCA on GPT~\cite{brown2020gpt} text embeddings as explained in Section~\ref{sec:zs}, we have to fit PCA on image embeddings using an internal database of 10.000 images~\footnote{https://tinyurl.com/kaggle-internal-dataset}.

We prepared $\times4$ \texttt{ViT-L-14} models and $\times5$ \texttt{ViT-H-14} using the previous fine-tuning strategy. At least one model from each was trained without augmentations. All models were trained for 1 epoch using the ``Google Landmarks 2020 Dataset"~\cite{google_landmark} and the ``Products-10K" dataset~\footnote{https://www.kaggle.com/competitions/products-10k/}. We use standard augmentations that include: horizontal flips, resizing, rotations, color variations, Cutout\cite{devries2017cutout}, and soft AutoAugment\cite{cubuk2018autoaugment}. We perform our experiments on two NVIDIA RTX 3090 GPU. 

Following robust fine-tuning best practices from Wortsman \emph{et al.}~\cite{wortsman2021robustwise, wortsman2022model}, we use ``model soup" and ensemble the weights of the zero-shot fine-tuned models. We ensure diversity in our two sets of models, and perform this ensemble technique to obtain a single \texttt{ViT-L-14} and \texttt{ViT-H-14} models.
These robustness and generalization improvements come at no additional computational cost during fine-tuning or inference~\cite{wortsman2021robustwise}. We show this comparison in Table~\ref{tab:lb2}.

\begin{figure}[t]
    \centering
    \includegraphics[width=\linewidth]{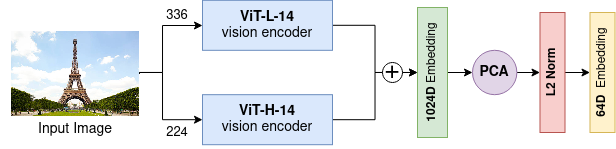}
    \caption{Our proposed \textbf{ensemble} of two ViT models~\cite{dosovitskiy2020vit, radford2021clip}.}
    \label{fig:ensemble}
\end{figure}

\paragraph{Final Solution: Ensemble}

Our solution is an ensemble of the two ViT models (from model soup \cite{wortsman2022model}), as previously explained each model produces a 512D descriptor. We created a $1024\mathcal{D}$ embedding using the concatenated outputs of both models. We then used PCA to reduce the dimension from 1024 to 64 and obtain the desired descriptor.
This process is illustrated in Figure \ref{fig:ensemble}. We found beneficial (empirically) to perform PCA on non-normalized vectors, and normalize after the PCA projection. In Table \ref{tab:lb2} we summarize the final experiments and ablations. Our final ensemble scored 0.688 $mP@5$ achieving \textbf{4th place} of 1022 participants in the challenge. We must note that contemporary challenge solutions also use some of these ideas.

\begin{table}[t]
    \centering
    \begin{tabular}{l|c}
         Method & Score \\
         \hline
         Baseline ZS \texttt{ViT-H-14} + PCA~\cite{dosovitskiy2020vit} & 0.603 \\
         \texttt{ViT-L-14} (single model)~\cite{dosovitskiy2020vit} & 0.657 \\
         \texttt{ViT-L-14} (model soup)~\cite{dosovitskiy2020vit, wortsman2021robustwise} & 0.669 \\
         \texttt{ViT-H-14} (model soup)~\cite{dosovitskiy2020vit, wortsman2021robustwise} & 0.677 \\
         Ensemble \texttt{ViT-L-14} + \texttt{ViT-H-14} & \textbf{0.688} \\
         Ensemble \texttt{ViT-H-14} + \texttt{ViT-H-14} & 0.682 \\
         \hline
    \end{tabular}
    \caption{Summary of experimental results on the leaderboard}
    \label{tab:lb2}
\end{table}

\paragraph{Implementation Details}

Our training routine consisted of taking a random sample of 9691 classes from Products-10k, and the Google Landmarks 2020 dataset\cite{google_landmark}. These 2 datasets represent roughly 50\% of the challenge data distribution. We only included classes which had at least 4 samples and limited to a maximum of 50 images per class.

We took several precautions in order to preserve the rich information encoded in the pre-trained weights of the CLIP model. We fine-tuned for a single epoch. We used \emph{layer wise learning rate decay}. This started at $1.25e^{-6}$ for the beginning layers of the model and climbed to a maximum of $10e^{-6}$ for the higher layers. We warmed up the model for 1000 steps and decreased the learning rate with a cosine schedule. This is important to keep the rich information encoded into the base ViT~\cite{dosovitskiy2020vit} models, and adapt it to the new data distribution.
The embedding layer (MLP-Head, see Figure \ref{fig:vit}) was trained with a higher learning rate $3e^{-4}$. We used a 512 MLP layer to act as our embedding layer; this had a multi-sample dropout added before being input into the ArcFace head, which consisted of sub-center ArcFace\cite{deng2019arcface} with adaptive margins; we set $c=3$ sub-centers.

We did not find any benefit in fine-tuning using a larger dataset, we believe adapting the model to a ``new" data distribution implies a more complex and longer fine-tuning. Also fine-tuning for more than a single epoch ``damaged" the original weights and zero-shot properties. 
We refer the reader to our public code for more details.
\url{https://github.com/IvanAer/G-Universal-CLIP}
%

\section{Conclusion}

The Google Universal Image Embedding (GUIE) Challenge opens the door to more general image retrieval methods and applications. Leveraging the information from zero-shot models pre-trained using CLIP at a large-scale is key. In this work we explore some beneficial properties from zero-shot models and robust fine-tuning tricks.

As future work we would like to investigate further the text-image descriptors alignment, and explore a similar approach for image compression.

\vspace{-2mm}

\paragraph{Acknowledgments}
Marcos Conde is supported by H2O.ai and the The Alexander von Humboldt Foundation.
We would like to thank Kaggle Team and Google Research for organizing the GUIE Challenge, especially: Andre Araujo, Francis Chen and Bingyi Cao.
We also thank OpenCLIP project and Ross Wightman for their brilliant open-source contributions that served as our baseline.

{\small
\bibliographystyle{ieee_fullname}
\bibliography{egbib}
}

\end{document}